%% file: egpaper_final.tex
\documentclass[10pt,twocolumn,letterpaper]{article}

\usepackage{cvpr}
\usepackage{times}
\usepackage{epsfig}
\usepackage{graphicx}
\usepackage{amsmath}
\usepackage{amssymb}

\usepackage[dvipsnames,table,xcdraw]{xcolor}
\usepackage{textcomp}
\usepackage{makecell}
\usepackage{rotating}
\usepackage{subfig}
\DeclareMathAlphabet{\pazocal}{OMS}{zplm}{m}{n}
\newcommand{\Image}{\pazocal{I}}
\newcommand{\LogImage}{\pazocal{G}}
\newcommand{\Irradiance}{\pazocal{E}}
\newcommand{\Lighting}{\boldsymbol{\pazocal{L}}}
\newcommand{\Attention}{\pazocal{A}}
\usepackage{textcomp}
\usepackage{mathtools}
\usepackage{booktabs}
\usepackage{tabularx}

\DeclareMathAlphabet{\pazocal}{OMS}{zplm}{m}{n}
\newcommand{\Loss}{\mathcal{L}}    

\newcommand{\Normal}{\pazocal{N}}
\newcommand{\Render}{\pazocal{R}}

\newcommand{\Light}{\boldsymbol{\ell}}
\newcommand{\Parameters}{\boldsymbol{\theta}}
\newcommand{\Model}{\boldsymbol{f}}
\newcommand{\BlendW}{\lambda_{blend}}
\newcommand{\degree}[1]{${#1}^o$}
\def\360{\degree{360}}

\usepackage{authblk}

\usepackage[pagebackref=true,breaklinks=true,letterpaper=true,colorlinks,bookmarks=false]{hyperref}

\hypersetup{
    colorlinks = true,
}
\cvprfinalcopy 

\def\cvprPaperID{17} 

\ifcvprfinal\fi
\usepackage[font=small]{caption}
\begin{document}
\title{\vspace{-1em}Deep Lighting Environment Map Estimation from Spherical Panoramas}

\author[1]{Vasileios Gkitsas\thanks{Equal contribution}}
\author[1,2]{Nikolaos Zioulis$^*$}
\author[2]{Federico Alvarez}
\author[1]{Dimitrios Zarpalas}
\author[1]{Petros Daras}
\affil[1]{\small Information Technologies Institute (ITI), Centre for Research and Technology Hellas (CERTH)}
\affil[2]{\small Signals, Systems and Radiocommunications Department (SSRD), Universidad Polit\'{e}cnica de Madrid (UPM)}

\maketitle
\thispagestyle{empty}

\begin{abstract}
Estimating a scene's lighting is a very important task when compositing synthetic content within real environments, with applications in mixed reality and post-production.
In this work we present a data-driven model that estimates an HDR lighting environment map from a single LDR monocular spherical panorama.
In addition to being a challenging and ill-posed problem, the lighting estimation task also suffers from a lack of facile illumination ground truth data, a fact that hinders the applicability of data-driven methods.
We approach this problem differently, exploiting the availability of surface geometry to employ image-based relighting as a data generator and supervision mechanism.
This relies on a global Lambertian assumption that helps us overcome issues related to pre-baked lighting.
We relight our training data and complement the model's supervision with a photometric loss, enabled by a differentiable image-based relighting technique.
Finally, since we predict spherical spectral coefficients, we show that by imposing a distribution prior on the predicted coefficients, we can greatly boost performance.
Code and models available at \href{https://vcl3d.github.io/DeepPanoramaLighting/}{vcl3d.github.io/DeepPanoramaLighting}
\end{abstract}

\section{Introduction}
\input{Introduction.tex}

\section{Related Work}
\label{sec:related}
\input{RelatedWork.tex}

\section{Lighting Estimation via Relighting}
\label{sec:method}
\input{Imaged-basedRelighting.tex}

\section{Results}
\label{sec:results}
\input{Results.tex}

\section{Discussion}
\label{sec:discussion}
\input{Discussion.tex}

\textbf{Acknowledgements:}
We acknowledge HW support by NVidia and financial support by the H2020 EC project Hyper360 (GA 761934).

{\small
\bibliographystyle{ieee_fullname}
\bibliography{./bibs/egbib,./bibs/data-driven,./bibs/datasets,./bibs/history}
}

\end{document}

%% file: Introduction.tex
\begin{figure}[!t]
\begin{center}
   \includegraphics[width=\columnwidth]{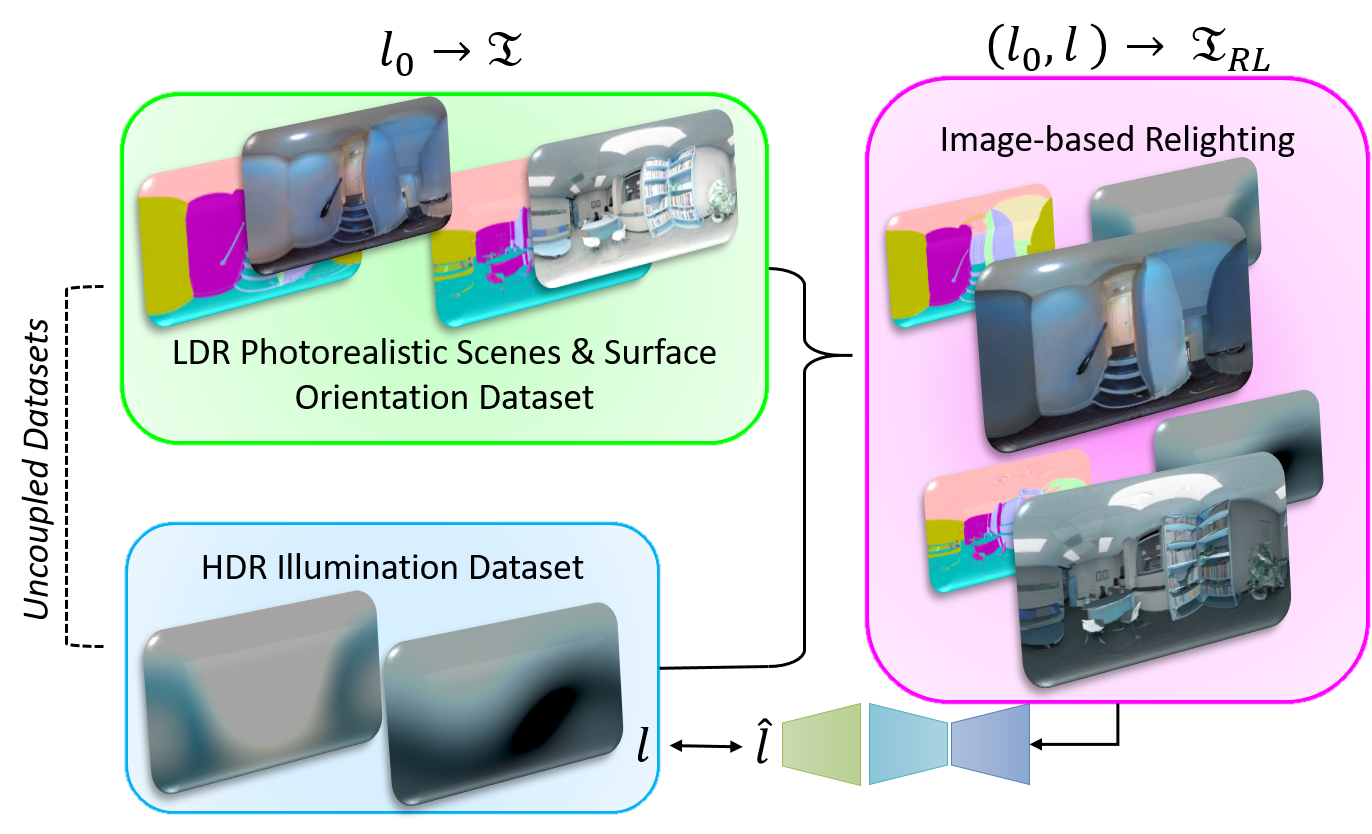}
\end{center}
   \caption{We use uncoupled datasets of HDR illumination maps \cite{gardner2017learning} and photorealistic spherical panoramas \cite{karakottas2019360} to train a spherical lighting environment map estimation model.
   While the former offer small content variance, they provide ground truth lighting $\Light$.
   While the latter offer multimodal data, they come with color content $\Image$ that has pre-baked lightin $\Light_o$.
   We combine these uncoupled datasets seamlessly using image-based relighting, producing relit images $\Image_{RL}$ and condition our supervision solely on the imposed lighting $\Light$. 
   }
\label{fig:concept}
\end{figure}

Compositing content from different domains (\textit{i.e.}~real and synthetic) has been an important part of post-production and visual effects.
It has now also become very relevant due to the maturation of mixed reality technologies.
These emerging technologies operate in between the real and virtual domains by embedding digitized or digital content in other media.

The realism of this composition depends on the positioning and lighting of the emplaced content, which is very demanding to accomplish, especially for monocular content as their estimation depends on solving ill-posed problems.
Data-driven methods have advanced the state-of-the-art in challenging tasks such as geometry estimation from single monocular images \cite{eigen2014depth}, and have even overcome the shortage of high quality data through self-supervision \cite{zhou2017unsupervised}.

Nonetheless, this is far more challenging for illumination estimation, due to the very complex process associated with the coloring of each pixel in an image.
Rarely -- if not ever -- do we see in the real world a scene to be exclusively lit from a single light source.  
In most cases illumination is composed of multiple sources including reflections, localized light sources (\textit{e.g.}~spot-lights) and/or broader area distributions (\textit{e.g.}~sunlight).

For the realistic relighting of embedded content, a high dynamic range (HDR) illumination needs to be estimated.
This makes data collection problematic and impractical as predictions will typically be done on a low dynamic range (LDR) image.
The Laval Indoor HDR Dataset \cite{gardner2017learning} only contains a very limited -- in terms of modern data-driven methods' requirements -- amount of coupled LDR and HDR spherical panoramas, capturing a scene's global lighting conditions, with some being duplicate captures \cite{weber2018learning}.
The Matterport3D dataset \cite{chang2018matterport3d}, used in \cite{song2019neural} to generate illumination environment maps, contains saturated HDR images and misses ceilings, where most of the indoor lighting comes from.
Further, the geometric warping used to register the LDR and HDR images resulted in high resolution artifacts.
Another direction \cite{garon2019fast, sengupta2018sfsnet, li2018learning} would be to use synthetic scenes, but given the lack of realistic materials and textures, the model's transferability to real world data would be limited.
Other approaches rely on specialized capturing with expensive hardware \cite{meka2019deep, sun2019single} or customized spherical objects \cite{legendre2019deeplight}.

Currently, the biggest challenge that data-driven illumination estimation methods need to overcome is the availability of data. 
Both \cite{meka2019deep} and \cite{sun2019single} only captured 18 subjects, while \cite{gardner2017learning} contains less than 2000 high quality LDR-HDR pairs.
At the same time, Matterport3D offers a large amount of data but does not offer high quality HDR environment maps, with the derivative dataset of \cite{song2019neural} coupling its generated HDR data with severely limited LDR samples.

Instead, we design a strategy that will exploit the strengths of all available datasets in order to train a model with the highest possible data variability -- in terms of photometric content and illuminations.
Our concept, as presented in Fig.~\ref{fig:concept}, is based on image re-lighting (RL).
We relight a single image $\Image$ with the lighting parameters $\Light$ through a rendering operation $\Render$:
\begin{equation}
\label{eq:relighting}
    \mathbf{\Image}_{RL} = \Render(\Image, \Normal, \Light),
\end{equation}
where $\Normal$ represent dense per pixel metadata.
Given estimated lighting parameters $\mathbf{\hat{\Light}} = \Model(\Parameters, \Image_{RL})$ predicted by a model $\Model$ with parameters $\Parameters$ using $\Image_{RL}$ as input, we can train a model with uncoupled color and ground truth illumination information as long as we condition our supervision on $\Light$ alone.
This is necessary as the image $\Image$ is a real world acquired image and has baked lighting into it.
It is therefore the result of another physically based rendering operation that involved the original scene's lighting parameters $\Light_o$.

This concept crucially relies on a global Lambertian assumption \cite{basri2003lambertian}. 
This widely used assumption \cite{cheng2018learning, sengupta2018sfsnet} is a practical approach to alleviate the complexity of natural illumination.
Under this assumption, we ensure that the original image $\Image$ -- with pre-baked lighting -- is a purely diffuse surface, effectively ignoring $\Light_o$.
In addition, under distant illumination, the relighting operation only depends on the surface's global direction.
As a result, our concept allows us to use high quality illumination maps (\textit{i.e.}~from \cite{gardner2017learning}) and exploit the content plurality and multimodality of modern datasets (\textit{i.e.}~\cite{chang2018matterport3d}) to train a global lighting map estimation model.
Summarizing, in this work we train a model to estimate a lighting environment map from a single monocular spherical panorama. 
Specifically, our contributions are:
\vspace{-5pt}
\begin{itemize}
    \item We use an efficient lighting representation and an image relighting rendering operation to synthesize randomized relit samples and thus, merge uncoupled datasets. 
    We complement this with environment map blending to increasing the variability of our training data.    
    \item As our image relighting operation is fully differentiable, we also leverage it on the supervision end to photometrically supervise our relit samples, enforcing a loss only on the imposed lighting.
    \item We further constrain our efficient spectral representation with a distribution prior that aids training offering a large performance gain.
\end{itemize}

%% file: RelatedWork.tex
Traditionally, lighting estimation is a long-standing problem which is very challenging as it requires geometrical (\textit{i.e.}~3D) data, camera (\textit{i.e.}~projection) characteristics, as well as material (\textit{i.e.}~attributes) metadata.
Early progress relied on the acquisition of 3D shapes \cite{marschner1997inverse} and camera intrinsics information in order to estimate a scene's lighting.
Nonetheless, even modern methods \cite{lombardi2015reflectance} rely on accurate geometry information or on assumptions about it \cite{lopez2013multiple}.
This line of work eventually matured into a single hypothesis estimation that explains the entirety of the scene (geometry and illumination) relying on natural scene statistics \cite{barron2014shape}.

The seminal work of \cite{Debevec:1998:RSO:280814.280864} first showed that it is possible to capture a scene's HDR lighting by capturing differently exposed images of a mirrored metallic sphere that reflects light.
Using such light probes has even enabled real-time video capture of HDR illumination at each position it is placed at within a scene.
Multiple spherical light probes have also been used to estimate a scene's global illumination \cite{debevec2004estimating}.
More recently, a specialized shading probe was designed for capturing directly the scene's shading for the purposes of mobile Augmented Reality (AR) \cite{calian2013shading}.
Nevertheless, the physical placement of known 3D objects into the scenes is not always practical as the images whose lighting needs to be estimated might have been priory captured.

Even before the advent of deep models, the intractability of the problem of lighting estimation lend itself to data-driven priors \cite{lalonde2009estimating} learned from millions of images.
Yet, modern data-driven methods are much more suited to overcoming the challenges of ill-posed tasks.
The main challenge that supervised data-driven methods need to overcome is the acquisition and/or estimation of supervisory data.
Aiming to estimate outdoor illumination in \cite{hold2017deep}, a synthetic sky model was fit to the captured panoramas in order to acquire the illumination parameters to regress.
To address the shortage of data, \cite{weber2018learning} harvested a set of light sources from the Laval Indoor HDR dataset and used them to augment the dataset's panoramas themselves, with the goal of learning to predict the illumination from monocular images of known 3D objects. 
This was achieved by aligning the latent space of HDR panoramas using an autoencoder, and then predicting the illumination map from a perspective render of it, using lighting from the dataset's panoramas in a supervised manner.

A custom data collection rig with three reflective spheres and a mobile phone was designed in \cite{legendre2019deeplight} to acquire a very large dataset of paired images with reflective spheres imaged with the corresponding illumination.
Then, an image-relighting supervision scheme was employed that relied on the photometric consistency between the sphere images collected by the capturing rig and the corresponding renders with the predicted illumination.
Generalization to indoor and outdoor scenes equally, enabled mobile mixed reality.

Synthetic content can be used to bypass cumbersome and custom data collection like in \cite{garon2019fast}, where the SunCG dataset \cite{song2017semantic} was used, which contains light positions.
After generating scenes with randomized lighting parameters, a set of random light probes were positioned into these scenes and subsequently rendered into RGB-D cubemaps.
A pre-trained network was used along with a discriminator  to allow for better inter-domain generalization.
It was trained to predict spatially varying light, conditioned by a specific location in each image.
This was achieved by the addition of an extra modality (\textit{i.e.}~depth) offered by the synthetic data and a dual path (\textit{i.e.}~global and local) network architecture.
Taking one step beyond, a parametric lighting representation was estimated in \cite{gardner2019deep}, with the goal being to more accurately model the localized nature of lighting.
Per light information (\textit{i.e.}~position, direction, color and size) was extracted from the Laval Indoor HDR dataset which was also used for providing the color data as near field of view crops of the panoramas.
Similar to \cite{garon2019fast}, a pre-trained encoder was used to alleviate the lack of data and fine-tuned to the lighting estimation task.
When used with geometry information, this parametric estimation allows for finer grained lighting results that respect visibility.

All the aforementioned methods estimate a scene's global (\textit{i.e.}omnidirectional) illumination from a single perspective image.
The recent work of \cite{song2019neural} addressed this through an image warping step, followed by an omnidirectional completion one, and then by a LDR to HDR estimation, finally encoding the HDR representation into an irradiance environment map. 
It was demonstrated that an end-to-end pipeline for these three tasks offered better illumination estimates.
The aforementioned work also contributed a novel dataset by sampling from perspective HDR images into omnidirectional HDR maps at specific locales.
In this way, their location conditioned predictions estimated localised spherical illumination from perspective images.
However, the Matterport3D HDR captures are satured and do not always include ceilings, meaning that important lighting information may be missed.
In addition, the geometric warping produces LDR and HDR images with severe high frequency artifacts which are not realistic.
Nonetheless, training on a higher variety of data and relying on omnidirectional supervision for the completion task helped in producing high quality omnidirectional lighting estimates, albeit object localised.

In our work, we estimate the illumination directly from a spherical panorama which captures global scene information more accurately.
We also use the original indoor scene images from Matterport3D instead of the distorted object localised ones.
This means that we need to overcome the lack of paired lighting information.

%% file: Imaged-basedRelighting.tex
Our concept is based on the works for precomputed radiance transfer \cite{sloan2002precomputed} and irradiance maps \cite{ramamoorthi2001efficient}.
These works use spherical harmonics (SH) representations to precompute intermediate function evaluations and use them during rendering.
SH are a frequency space representation\footnote{While generally defined on imaginary numbers, we will only be referring to real valued functions, and thus, our definition of SH corresponds to Real Spherical Harmonic functions.} of a function over the sphere, analogous to the Fourier transform.
More details about SH in general and SH lighting can be found in \cite{green2003spherical}, however, we will followingly present the necessary preliminaries.

\subsection{Spherical Harmonics Diffuse Lighting}
Given that SH form an orthonormal basis over the sphere, each spherical function can be represented as a linear combination of a set of basis function:
\begin{equation}
\label{eq:sh_reconstruction}
    \hat{\Lighting}(\mathbf{\rho}) = \sum_{\mathbf{\rho}}^{\Omega} L_m^l Y_m^l(\mathbf{\rho}),
\end{equation}
where $\mathbf{\rho} = (\phi, \theta)$ are the spherical angular coordinates defined in the spherical image domain $\Omega$, $L_m^l$ is the coefficient of the $Y_m^l$ SH basis function of degree (\textit{i.e.}~band) $l$ and order $m$, spanning $-l \leq m \leq l$.
The function $\hat{\Lighting}$ is an approximation that depends on the depth of the order $l$, with smaller orders corresponding to lower frequencies.
In our context, the approximated function $\hat{\Lighting}$ represents the distant illumination.
The SH coefficients $L_m^l$ are calculated by 
\begin{equation}
\label{eq:sh_projection}
    L_m^l = \frac{4\pi}{N} \sum_{\mathbf{\rho}}^{\Omega} \Lighting(\mathbf{\rho}) Y_m^l(\mathbf{\rho})
\end{equation}
with $N = w \times h$ being the total discrete elements sampled, corresponding to the area of the spherical image domain of width $w$ and height $h$.

As shown in \cite{ramamoorthi2001efficient}, only the first $3$ orders (\textit{i.e.}~$l=[0,1,2]$, meaning $9$ coefficients) are needed to accurately represent the surface's irradiance (up to $99.2\%$ \cite{basri2003lambertian}), given that irradiance smoothly varies with orientation.
Consequently, SH coefficients are a very effective technique to compress direct illumination from distant sources.
For the remainder of this document we will consider SH coefficients only up to the third order $0 \leq \Light \leq 2$.
As presented in \cite{ramamoorthi2001efficient}, the resulting irradiance $\Irradiance$ map can be expressed as a function of the surface's orientation, represented by the normal map $\Normal$, and the SH coefficients $\Light = \{L_m^l\}$, after lifting them into a symmetric $4 \times 4$ matrix representation:
\setlength{\arraycolsep}{2pt}
\begin{equation}
\label{eq:irradiance}
    \Irradiance(\Normal, \Light) \! = \! \eta(\Normal)^T  \! \begin{bsmallmatrix}
        c_1 L_2^2 & c_1 L^2_{-2} & c_1 L^2_1 & c_2 L^1_1 \\
        c_1 L^2_{-2} & -c_1 L^2_2 & c_1 L^2_{-1} & c_2 L^1_{-1} \\
        c_1 L^2_1 & c_1 L^2_{-1} & c_3 L^2_0 & c_2 L^1_0 \\
        c_2 L^1_1 & c_2 L^1_{-1} & c_2 L^1_0 & c_4 L^0_0 - c_5 L^2_0
    \end{bsmallmatrix} \! \eta(\Normal),
\end{equation}
with $\eta(\Normal) = (\Normal_x, \Normal_y, \Normal_z, 1)$ being a homogeneous coordinates transformation operation.
The symmetric matrix and constants $c_i, i \in [1, 4]$ are the result of expanding the rendering equation and associated spherical harmonic normalization.
Overall, we find that SH lighting -- Eq. (\ref{eq:irradiance}) -- relies on a matrix-vector and a vector-vector product, which is fully differentiable, as well as parallelizable and thus, very well suited for modern tensor-based data-driven methods.
Also, the same applies to the SH projection -- Eq. (\ref{eq:sh_projection}) -- and reconstruction -- Eq. (\ref{eq:sh_reconstruction}) -- functions.

\subsection{Training with Uncoupled Datasets}
\begin{figure}[!htbp]
\begin{center}
   \includegraphics[width=\linewidth]{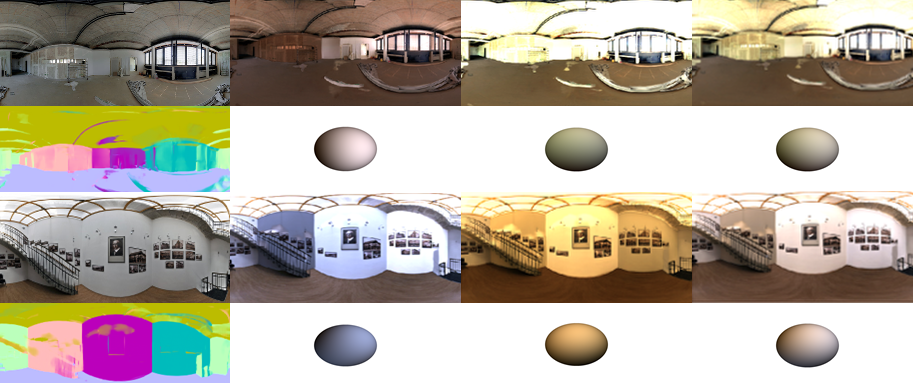}
\end{center}
   \caption{The combined effect of blending 2 lighting maps with an even ratio.
   On the left is the original color image and associated normal map.
   Then, on the right, 2 different lighting probes follow with their rendered lighting on the original image on top.
   Finally on the rightmost, we show the blended lighting probe as well as the rendered original image with the combined lighting on top.
   Lighting blending increases the variety of the dataset and can also help in overcoming non realistic relights.}
\label{fig:blending}
\end{figure}

While works like \cite{gardner2017learning} train on a limited amount of HDR panoramas, our goal is to learn illumination information from LDR images, similar to more recent works \cite{legendre2019deeplight, song2019neural, gardner2019deep, garon2019fast}.
Nonetheless, we need to regress HDR illumination in order to be able to convincingly add synthetic objects into natural scenes.
More recent datasets either do not offer HDR environment maps \cite{legendre2019deeplight} or their data come with deficiencies \cite{song2019neural}.
Seeking to avoid the problems associated to synthetic data \cite{garon2019fast}, instead, our approach is to blend different lighting conditions to randomly synthesize illumination maps.
Considering two HDR images representing dense lighting conditions $\Lighting_a$ and $\Lighting_b$ we can blend them using a ratio $\BlendW \in [0, 1]$ to produce a blended lighting map $\Lighting_{ab} = \BlendW \Lighting_a + (1 - \BlendW) \Lighting_b$. 
Then, using Eq. (\ref{eq:sh_projection}) we can project $\Lighting_{ab}$ to a set of SH coefficients $\Light_{ab} = \{L_m^l\}$ up to a predefined order $l$.
A set of examples of blending 2 lighting maps and their combined relighting effect in contrast to the original relights are presented in Fig.~\ref{fig:blending}.

Considering a spherical image $\Image$ and its corresponding, aligned normal map $\Normal$ we can relight the image $\Image$ to produce a new relit one $\Image_{RL}$ with a set of low dimensional SH lighting parameters $\Light$ using Eq.~(\ref{eq:irradiance}).
Accordingly, the rendering operation $\Render$ defined in Eq.~(\ref{eq:relighting}) for each pixel becomes:
\begin{equation}
\label{eq:sh_rendering}
    \Render(\Image, \Normal, \Light) := \Irradiance(\Normal, \Light) \,  \Image.
\end{equation}
Therefore, we can render new images $\Image_{RL} = \Render(\Image, \Normal, \Light)$ coupled with the newly imposed lighting conditions $\Light$ that will serve as our training data. 
Even though the original image contains pre-baked lighting, as presented in Fig.~\ref{fig:relit}, the new lighting can greatly influence the resulting image and change its overall appearance (through the ambient term), as well as relative scene luminance and shading. 
As long as we properly regularize our supervision on this newly imposed lighting alone, a lighting estimation model can be trained under this scheme.

This allows us to exploit larger photorealistic datasets with higher content variance, that additionally offer surface orientation information, and simultaneously exploit the higher quality, but limited in terms of scale, HDR illumination datasets.
The scale limitation is bypassed through the aforementioned blending process that enables the on-the-fly generation of a larger HDR illumination dataset with higher variability compared to approaches such as \cite{gardner2019deep} that only use the dataset's original lighting.
More specifically, we blend HDR illumination maps from the Laval Indoor HDR dataset \cite{gardner2017learning} and relight the realistic panoramas \cite{karakottas2019360} that contains raytraced normal maps as well.

\subsection{Deep Spherical Lighting Estimation Model}
\begin{figure*}[!t]
\begin{center}
   \includegraphics[width=\linewidth]{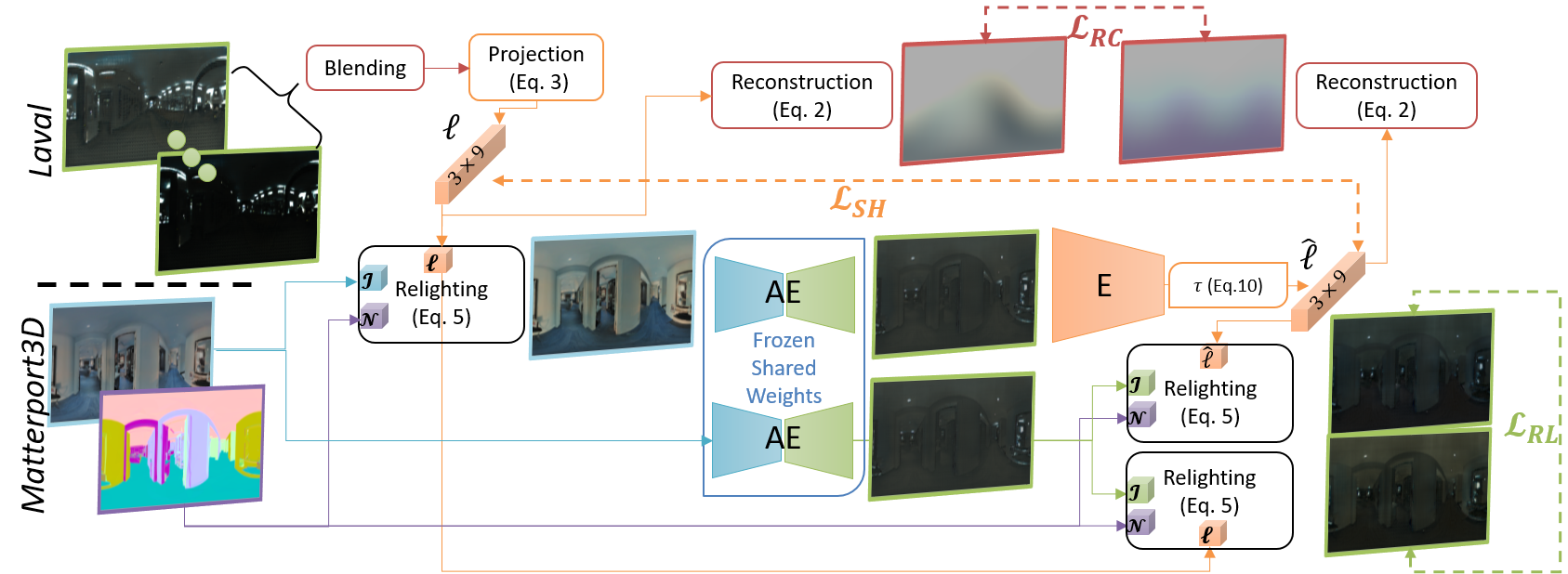}
\end{center}
   \caption{
Our end-to-end relighting based supervision. 
An input LDR spherical panorama image $\Image^{LDR}$ is along with its corresponding normal map $\Normal$ and randomly blended lighting parameters $\Light$, in the form of SH coefficients, is relit using the rendering operation $\Render$ of Eq.~(\ref{eq:sh_rendering}).
The relit LDR image $\Image^{LDR}_{RL}$ is translated to the HDR image $\Image^{HDR}_{RL}$ through the pre-trained (frozen weights) LDR-to-HDR autoencoder ($\textbf{AE}$).
The result gets encoded by a lighting encoder $\textbf{E}$ to the estimated lighting $\hat{\Light}$ which is passed through the spectral prior function $\tau$ (Eq.~\ref{eq:prior}).
The original LDR image also gets translated to the HDR domain, producing $\hat{\Image}^{HDR}$.
This gets rendered using using the aligned normal map $\Normal$ two times, once with the original light $\Light$ and once with the predicted light $\hat{\Light}$.
This way we synthesize $\hat{\Image}^{HDR}_{GT}$ and $\hat{\Image}^{HDR}_{RL}$ respectively which are used to photometrically supervise our lighting estimator ($\Loss_{RL}$).
In addition, the ground truth and predicted SH coefficients are used to reconstruct lighting environments maps $\Lighting$ and $\hat{\Lighting}$ respectively through Eq.~(\ref{eq:sh_reconstruction}).
Additional losses are defined for the sparse SH coefficients ($\Loss_{SH}$) and the dense lighting map reconstructions ($\Loss_{RC}$).
LDR images are with cyan border, HDR images and lighting parameters are with lime borders, while their low frequency reconstructions are with red, and normal maps are with violet borders.
}
\label{fig:model}
\end{figure*}

Our CNN model is trained to regress a compressed representation of a lighting environment map that can be used to evaluate the irradiance of elements rendered within a scene.
The input to our model is a LDR panorama in equirectangular format representing a scene for which we want to estimate the lighting.
Two sub-networks are combined in an end-to-end manner to achieve this as depicted in Fig.~\ref{fig:model}.
Since we need to regress to HDR illumination, a LDR-to-HDR autoencoder ($\textbf{AE}$) CNN is first used to translate the LDR image to the HDR domain.
We then use an encoder ($\textbf{E}$) to regress to the SH coefficients $\hat{\Light} \in \mathbb{R}^{9 \times 3}$ that represent a low frequency distant illumination ($9$ SH coefficients) for each of the $3$ color channels (\textit{i.e.}~red, green, blue).

\input{figures/Relighting.tex}

Each training sample comprises a color image $\Image^{LDR} \in \mathbb{R}^{W \times H \times 3}$, a normal map $\Normal \in \mathbb{R}^{W \times H \times 3}$, and the lighting coefficients $\Light$ calculated after randomly sampling two HDR illumination maps and blending them, to eventually produce the network's input which is a relit image $\Image_{RL}^{LDR}$, using Eq. (\ref{eq:sh_rendering}).
In this way, we re-use and couple distinct, uncoupled data of color images supported by surface information, with other data of illumination maps.
The relit LDR images gets translated by the LDR-to-HDR sub-network to an HDR image $\Image^{HDR}_{RL} \in \mathbb{R}^{W \times H \times 3}$ which then gets encoded to the predicted SH coefficients $\hat{\Light}$.

The SH representation and the availability of surface information allows our supervision to be multi-faceted.
First, we employ an L2 regression objective to the predicted coefficients, the coefficients loss:
\begin{equation}
\label{eq:coefficient_loss}
    \Loss_{SH} = \frac{1}{9 \times 3} \sum_{c=1}^3 \sum_{l=0}^2 \sum_{m=-2}^{2} \big(L_{m,c}^l - \hat{L}_{m,c}^l\big)^2
\end{equation}

In addition, through Eq.~(\ref{eq:sh_reconstruction}) we can reconstruct the lighting maps $\Lighting$ and $\hat{\Lighting}$ represented by the ground truth and estimated SH coefficients respectively.
As a result, we further regularize our supervision with a denser L2 objective, the reconstruction loss:
\begin{equation}
\label{eq:reconstruction_loss}
    \Loss_{RC} = \frac{1}{w \times h} \sum_{\mathbf{\rho}}^{\Omega} \Attention(\mathbf{\rho}) \big\| \Lighting(\mathbf{\rho}) - \hat{\Lighting(\mathbf{\rho})} \big\|^2_2,
\end{equation}
where $\Attention \in \mathbb{R}^{W \times H \times 3}$ is the spherical attention mask used in \cite{zioulis2019spherical}.
In this way, pixels towards the equator are weighted more appropriately with respect to those near the poles that are sampled multiple times due to the equirectangular distortion and whose spherical area is smaller.

Finally, we can also compute a dense photometric consistency loss \cite{godard2017unsupervised} between an image relit with the original lighting and that when relit using the model's regressed output.
As depicted in Fig~\ref{fig:model}, we convert the original (non-relit) LDR image to HDR, producing $\hat{\Image}^{HDR}$.
Using Eq.~(\ref{eq:sh_rendering}) and the normal map $\Normal$, we render this HDR image  with the ground truth SH coefficients $\Light$, producing $\hat{\Image}^{HDR}_{GT}$, and with the estimated SH coefficients $\hat{\Light}$, creating $\hat{\Image}^{HDR}_{RL}$.
Our relighting loss is then formulated as:
\begin{equation}
\label{eq:photometric_loss}
    \Loss_{RL} \! = \! \frac{1}{N} \! \sum_{\mathbf{\rho}}^{\Omega} \! \Attention \big( \alpha | \hat{\LogImage}_{GT} \! - \! \hat{\LogImage}_{RL} | \! + \! (1 - \alpha) SD(\hat{\LogImage}_{GT}, \hat{\LogImage}_{RL}) \big),
\end{equation}
where $SD$ is the structural dissimilarity function, $\alpha$ a blending factor between $SD$ and the L1 loss, while $\LogImage$ is the $\log$ transformed HDR image (superscripts and $\mathbf{\rho}$ indexing omitted for clarity).
The L1 term in the log domain allows us to penalize big relighting errors, while the structural dissimilarity term penalizes shading discrepancies within regions of the image.

Therefore, our final weighted objective is:
\begin{equation}
\label{eq:total_loss}
    \Loss = \lambda_{SH} \Loss_{SH} + \lambda_{RC} \Loss_{RC} + \lambda_{RL} \Loss_{RL}
\end{equation}
It should be noted that relighting the LDR image requires a scaling of the SH coefficients to produce realistic and properly saturated relights. 
However, this is only needed for the input LDR alone, in order to modify its lighting.
Supervision, regression and photometric alike, is facilitated by the unscaled -- HDR -- SH coefficients, allowing the network to learn HDR lighting from a single LDR image.

All prior works that regress SH coefficients directly operate on the model regressed values.
In this way, the predicted values are in an unstructured form driven purely by the supervision.
However, SH coefficients are spectral domain values and follow a structured distribution.
The DC term (\textit{i.e.}~ambient or $L_0^0$) is the strongest coefficient, with the magnitude of the other frequencies dropping according to their order (\textit{i.e.}~$L_{\{-1,0,1\}}^1 >> L_{\{-2, -1, 0, 1, 2\}}^2$).
We use this spectral distribution prior to enforce a structured SH prediction on each channel's predicted coefficient vector $\hat{\Light}$ by applying a function $\tau(\hat{\Light})$, where:
\begin{equation}
\label{eq:prior}
    \tau(\Light) = \| \Light \| \sigma(\Light).
\end{equation}
The normalized exponential function $\sigma$ (\textit{i.e.}~softmax) pushes lower values lower, and higher values higher, therefore easing the structuring of our predictions and aiding the supervision.
Since it also normalizes to unity, we first extract the magnitude of the predicted coefficient vector, and reapply it after enforcing the distribution prior.

Finally, both the scaled coefficients used to relight the LDR input image, as well as the predicted coefficients are deringed.
Ringing, also called Gibbs Phenomenon, was analyzed in \cite{sloan2008stupid} and relates to negative values during rasterization with the SH coefficients.
Thus, we reduce ringing artifacts by multiplying the SH coefficients with a low pass filter \cite{sloan2017deringing}, enforcing the coefficients per band to decrease smoothly to zero at some cut-off frequency.
Our model is end-to-end trainable as all operations are differentiable.

%% file: figures/Relighting.tex
\begin{figure}[!tbp]
    \centering
\subfloat[]{\includegraphics[width=0.24\columnwidth]{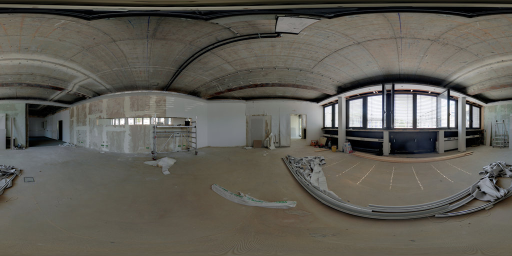}} \!
\subfloat[]{\includegraphics[width=0.24\columnwidth]{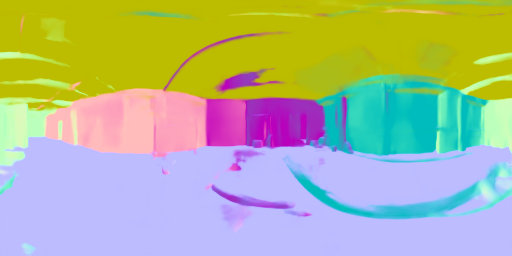}} \!
\subfloat[]{\includegraphics[width=0.24\columnwidth]{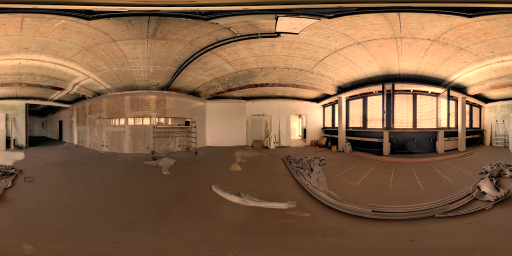}} \!
\subfloat[]{\includegraphics[width=0.24\columnwidth]{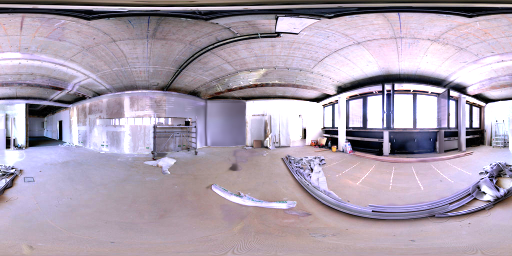}} \\[-5ex]
\subfloat[]{\includegraphics[width=0.24\columnwidth]{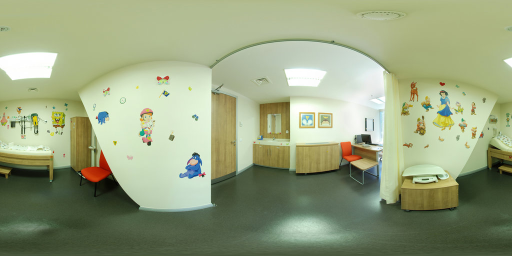}} \! 
\subfloat[]{\includegraphics[width=0.24\columnwidth]{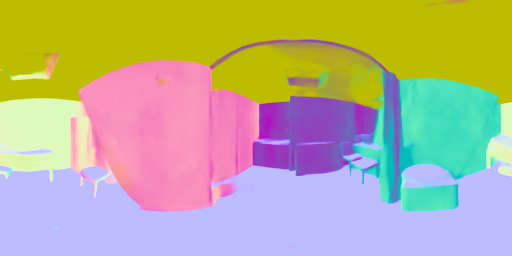}} \! 
\subfloat[]{\includegraphics[width=0.24\columnwidth]{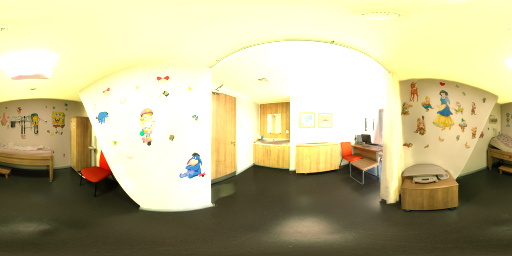}} \! 
\subfloat[]{\includegraphics[width=0.24\columnwidth]{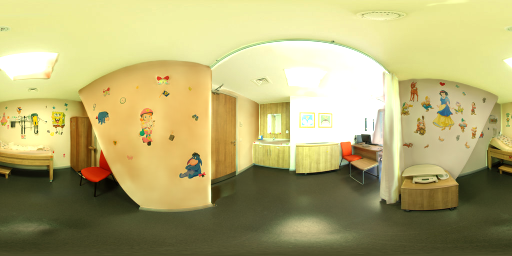}} \\[-5ex] 
\subfloat[Original Image]{\includegraphics[width=0.24\columnwidth]{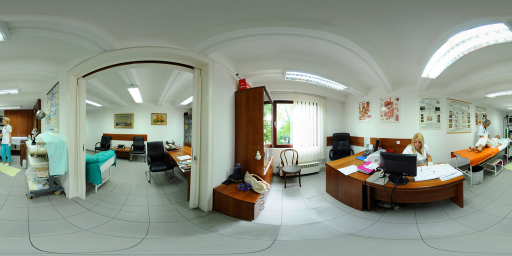}} \! 
\subfloat[Normal Map]{\includegraphics[width=0.24\columnwidth]{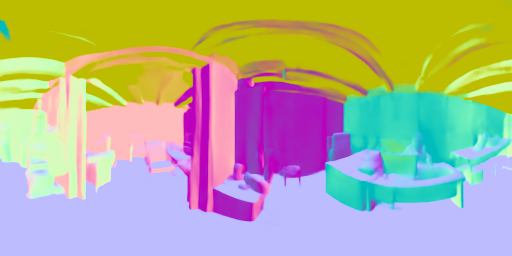}} \! 
\subfloat[Relit \#1]{\includegraphics[width=0.24\columnwidth]{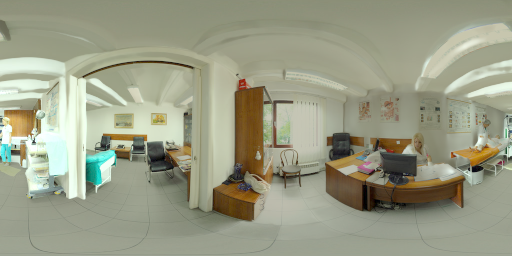}} \! 
\subfloat[Relit \#2]{\includegraphics[width=0.24\columnwidth]{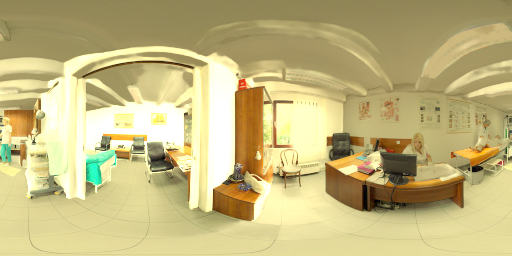}} 
    \caption{Exemplary relights of sample images, presented with their corresponding normal maps.
    The relit images' lighting is highly modified and draws away from the original lighting.
    This allows us to largely overcome the pre-baked lighting while training, by considering the original image as a diffuse surface.
    Thus, we complement our supervision with a photometric supervision relighting loss that is only conditioned on the newly imposed light.}
    \label{fig:relit}
\end{figure}

%% file: Results.tex
\subsection{Implementation Details}
Our experiments were run on a single machine with an i7 processor and a NVidia Titan X GPU, using PyTorch \cite{NIPS2019_9015}. 
All the models were trained for $10$ epochs using the Adam optimizer \cite{kingma2014adam} with its default parameters and an initial learning rate of $1\times 10^{-4}$.
The resolution $W \times H$ of the color images and their corresponding normal map is is $512 \times 256$.
We generate randomized lighting coefficients from the Laval HDR Indoor Dataset \cite{gardner2017learning}, after blending the dataset's HDR images in their original resolution and projecting them to SH using Eq.~(\ref{eq:sh_projection}).
However, when reconstructing them for the evaluation of $\Loss_{RC}$ using Eq.~(\ref{eq:sh_reconstruction}) we reconstruct them in the training images' resolution.
Out of the $2235$ images of the Laval dataset, we used $1682$ for training and out of the remaining images, we kept $443$ for evaluation purposes, while skipping $110$ images due to very low brightness that led to very dark relit images.
When relighting the LDR images, we scale the SH coefficients by $100$ to align the dynamic ranges appropriately for LDR inputs.

The LDR-to-HDR autoencoder's architecture was inspired by \cite{hdrcnnEilertsen2017HDRIR}.
We use the dataset of \cite{song2019neural} to train this sub-network, after resizing its coupled LDR and HDR images to $512 \times 256$, and directly supervise it with a $L2$ reconstruction loss on the predicted HDR, and a $L2$ reconstruction loss on the reconstructed low frequency lighting environment map, which is  the result of projecting the predicted HDR image to the SH basis and then reconstructing it from these SH coefficients. 
We use a learning rate of $1\times 10^{-3}$ and the Adam optimizer with its default parameters.
We train this network for $30$ epochs without any weighting between the 2 losses.
After convergence, we freeze the weights of the LDR-to-HDR model and use it to translate the input images to the HDR domain when training our lighting estimation encoder.

For the HDR lighting encoder, we base our model on the corresponding architecture of \cite{hold2017deep}, using $7$ convolutional and $2$ fully connected layers with ELU activation functions except for the head of the encoder that regresses the output coefficients.
When blending pairs of HDR environment maps from the Laval dataset, we randomly (uniform) sample two lighting maps and blend them using $\lambda_{blend} = 0.5$.
This allows us to greatly increase the diversity of the estimated lightings.
Our loss weights are $\lambda_{SH} \! = \! 0.01$, $\lambda_{RC} \! = \! 0.3$ and $\lambda_{RL} \! = \! 0.7$.

\subsection{Quantitative Comparisons}
Up to now, the state-of-the-art has focused on researching illumination estimation methods for traditional perspective images.
On the other hand, our method estimates global lighting from omnidirectional images.
Consequently, direct comparison is feasible with methods that do not rely on similar assumptions or learned perspective priors.
We offer comparison results to SIRFS \cite{barron2014shape} which represents the state-of-the-art in learning free illumination estimation.
We report the median scaled RMSE (m-RMSE) for the predicted lighting environment maps and the ground truth ones.
Median scaling uses the median of the evaluated and ground truth signals to scale the former to the value range of the latter.
In this way we offer more meaningful comparisons between different methods by removing any scale ambiguity between their predictions which is required for numerous reasons.
Methods like SIRFS that analytically solve the problem using LDR inputs, cannot easily regress to the correct HDR scale.
Also, different datasets might offer HDR images at different scales, a prominent example being the saturated HDRs of Matterport3D.
Finally, tone mapping, gamma correction, exposure times and color calibration are all different issues that influence the final compositing result.
In the end, numeric accuracy might not always translate to high quality compositing, but instead, relatively correct estimates that better capture the light's orientation and drop-off will surely produce more visually pleasing results.
Finally, the reported m-RMSE uses spherical distortion weighting when evaluating the MSE, similar to \cite{zioulis2019spherical}, in order to reduce the metric skewing effect that equirectangular distortion introduces.

Table~\ref{tab:results} presents the median scaled RMSE for our model and SIRFS on the test set of the Laval Indoor HDR dataset.
The ground truth is generated by projecting the HDR images into the spherical harmonics basis using Eq.~(\ref{eq:sh_projection}) and then reconstruct the ground truth lighting environment maps using Eq.~(\ref{eq:sh_reconstruction}).
As both SIRFS and our models estimate SH coefficients, we reconstruct the corresponding dense environment maps and calculate m-RSME on these.
Evidently, our model outperforms SIRFS, but more importantly it should be noted that the Laval images are \textit{unseen} for our model, and that the estimated lighting maps never participated entirely during training due to blending.

Table~\ref{tab:results} also presents results for a set of ablation experiments.
Overall, we observe a close to $50\%$ performance boost when imposing the spectral prior function $\tau$ on our predictions as indicated by the middle entries (without the prior), compared to the bottom entries (with the prior).
Interestingly, relying purely on the relighting loss (the \textit{only photo} variant with $\lambda_{SH} \! = \! 0, \lambda_{RC} \! = \! 0$) allows for proper model training.
Further, training without the relighting loss (the \textit{no photo} variant with $\lambda_{RL} \! = \! 0$) hurts performance, indicating its significance. 
At the same time, relying purely on the dense reconstructed signal loss (the \textit{only dense} variant with $\lambda_{SH} \! = \! 0, \lambda_{RL} \! = \! 0$) showcases inferior performance even to SIRFS.
More importantly, the discrepancy is a lot larger when not using the spectral prior function $\tau$, whilst with it, its implicit regularization does not allow for significant performance reduction.
\input{tables/results.tex}

\subsection{Qualitative Results}
\input{figures/Qualitative.tex}
Finally, we present qualitative results in Fig.~\ref{fig:qualitative} of synthetic object rendering into the spherical panoramas using our estimated lighting environment maps.
We use the Mitsuba \cite{jakob2010mitsuba} ray-tracer engine and the Bunny and Armadillo models from the Stanford 3D dataset \cite{levoy2005stanford}.
Each model is rendered with $512$ samples and is presented with three different materials.
We observe that the $0^{th}$ order SH coefficient $L_0^0$ -- the ambient lighting -- is well estimated, as well as the $1_{st}$ order $\{ L_{-1}^1, L_0^1, L^1_1 \}$ coefficients that encode directional lighting. 
Overall, our approach attains reasonable results even un completely unseen data harvested from the internet (first seven rows).

%% file: tables/results.tex
\begin{table}[]
\centering
\caption{
Median scaled RMSE results for the Laval test set.
}
\label{tab:results}
\begin{tabular}{@{}l|c|c@{}}
\toprule
\textbf{Method}   & \textbf{Prior}              & \textbf{m-RMSE} \\ \midrule
SIRFS             & No                          & 0.0391         \\ \midrule
Ours (\textit{full})       & \cellcolor[HTML]{D9E5F6}No  & 0.0229          \\
Ours (\textit{only photo}) & \cellcolor[HTML]{D9E5F6}No  & 0.0297          \\
Ours (\textit{no photo})   & \cellcolor[HTML]{D9E5F6}No  & 0.0334          \\
Ours (\textit{only dense}) & \cellcolor[HTML]{D9E5F6}No  & 0.0423          \\ \midrule
Ours (\textit{full})       & \cellcolor[HTML]{D9F2DE}Yes & \textbf{0.0101} \\
Ours (\textit{only photo}) & \cellcolor[HTML]{D9F2DE}Yes & 0.0116          \\
Ours (\textit{no photo})   & \cellcolor[HTML]{D9F2DE}Yes & 0.0156          \\
Ours (\textit{only dense}) & \cellcolor[HTML]{D9F2DE}Yes & 0.0162          \\ \bottomrule
\end{tabular}
\end{table}

%% file: figures/Qualitative.tex
\begin{figure*}[!htp]
    \centering
\subfloat{\includegraphics[width=0.24\linewidth]{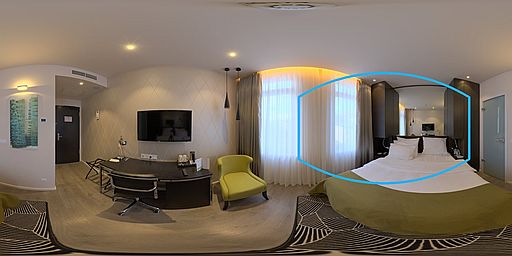}} 
\hspace{0.01 cm}
\subfloat{\includegraphics[width=0.24\linewidth]{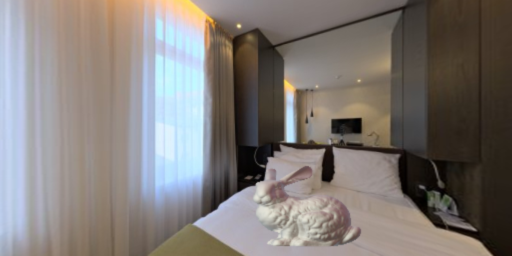}}
\hspace{0.01 cm}
\subfloat{\includegraphics[width=0.24\linewidth]{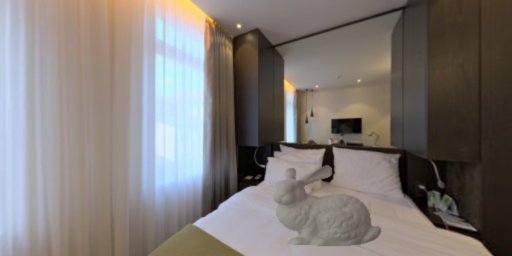}} 
\hspace{0.01 cm}
\subfloat{\includegraphics[width=0.24\linewidth]{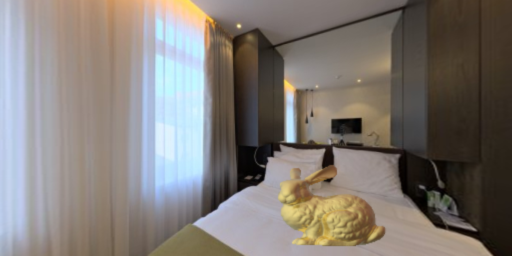}}

\subfloat{\includegraphics[width=0.24\linewidth]{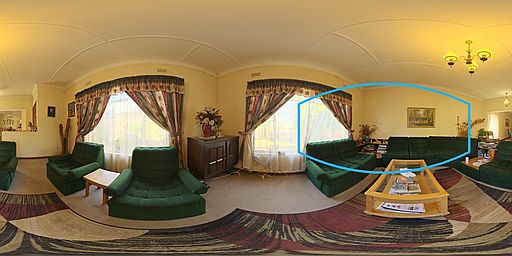}} 
\hspace{0.01 cm}
\subfloat{\includegraphics[width=0.24\linewidth]{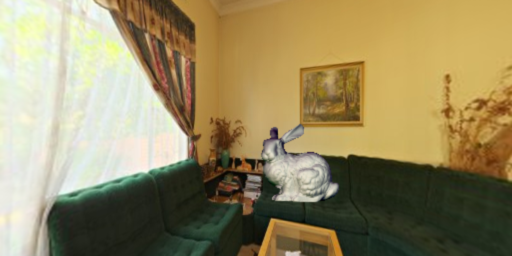}}
\hspace{0.01 cm}
\subfloat{\includegraphics[width=0.24\linewidth]{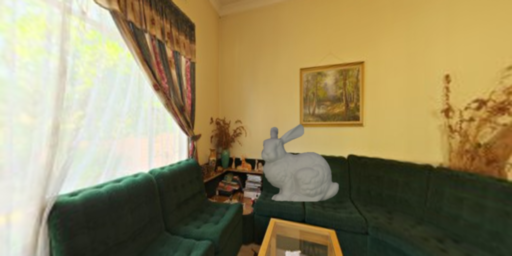}}
\hspace{0.01 cm}
\subfloat{\includegraphics[width=0.24\linewidth]{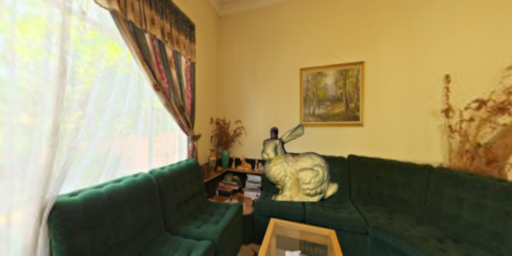}}

\subfloat{\includegraphics[width=0.24\linewidth]{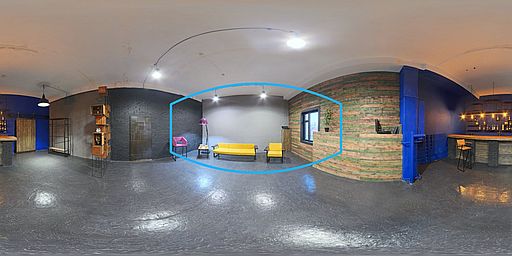}} 
\hspace{0.01 cm}
\subfloat{\includegraphics[width=0.24\linewidth]{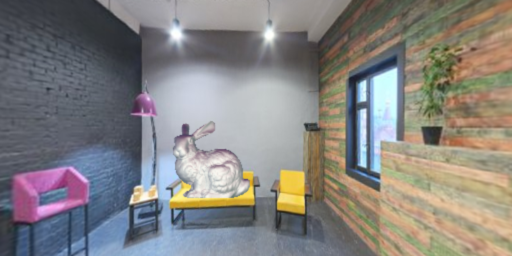}} 
\hspace{0.01 cm}
\subfloat{\includegraphics[width=0.24\linewidth]{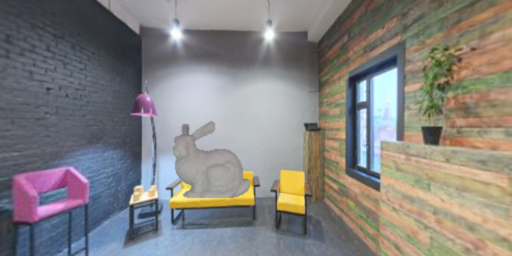}} 
\hspace{0.01 cm}
\subfloat{\includegraphics[width=0.24\linewidth]{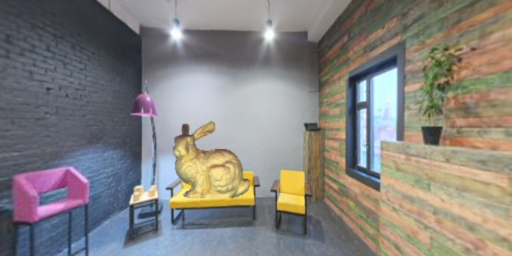}}

\subfloat{\includegraphics[width=0.24\linewidth]{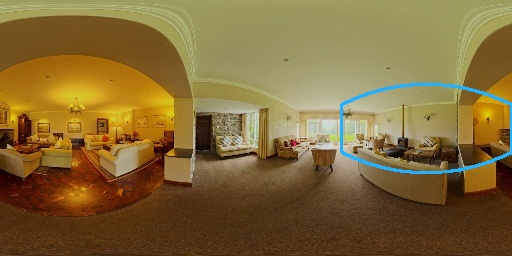}} 
\hspace{0.01 cm}
\subfloat{\includegraphics[width=0.24\linewidth]{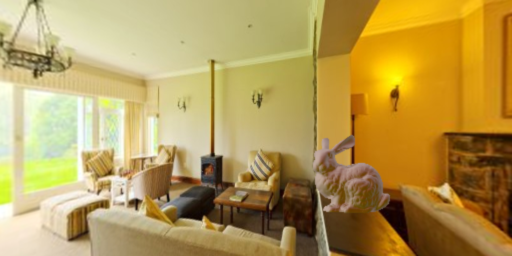}} 
\hspace{0.01 cm}
\subfloat{\includegraphics[width=0.24\linewidth]{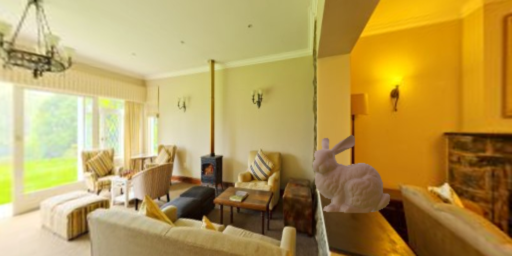}} 
\hspace{0.01 cm}
\subfloat{\includegraphics[width=0.24\linewidth]{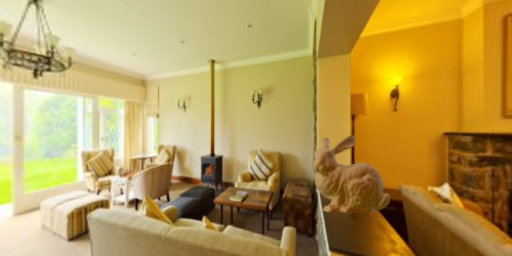}} 

\subfloat{\includegraphics[width=0.24\linewidth]{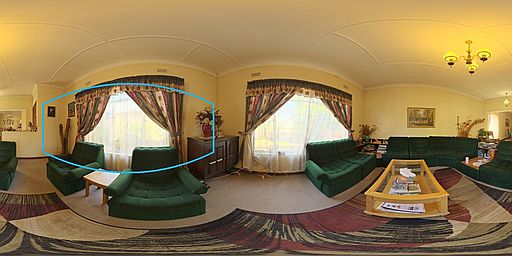}} 
\hspace{0.01 cm}
\subfloat{\includegraphics[width=0.24\linewidth]{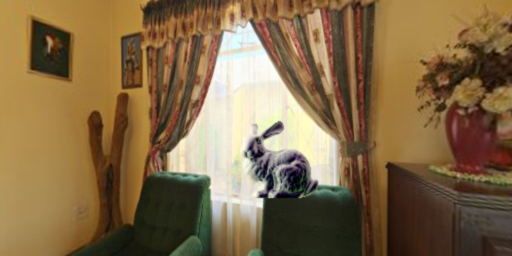}} 
\hspace{0.01 cm}
\subfloat{\includegraphics[width=0.24\linewidth]{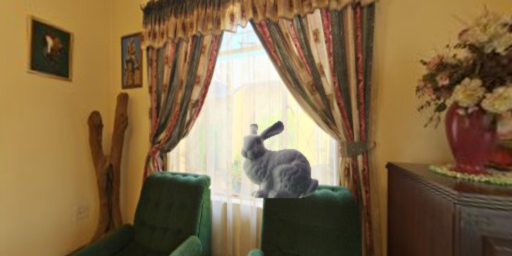}}
\hspace{0.01 cm}
\subfloat{\includegraphics[width=0.24\linewidth]{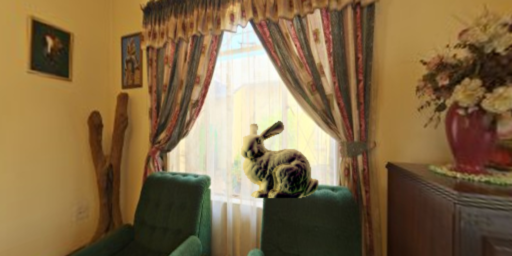}}

\subfloat{\includegraphics[width=0.24\linewidth]{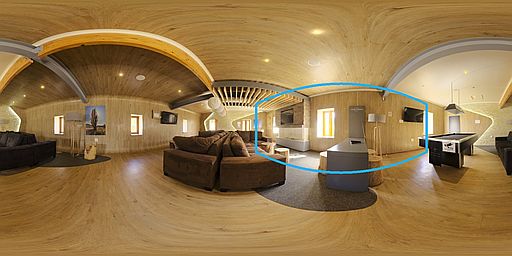}} 
\hspace{0.01 cm}
\subfloat{\includegraphics[width=0.24\linewidth]{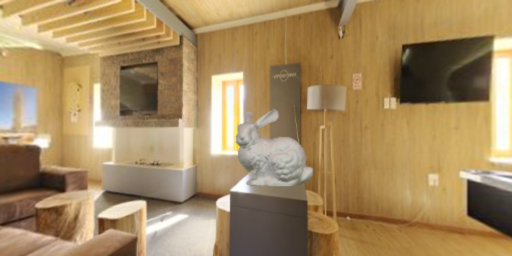}} 
\hspace{0.01 cm}
\subfloat{\includegraphics[width=0.24\linewidth]{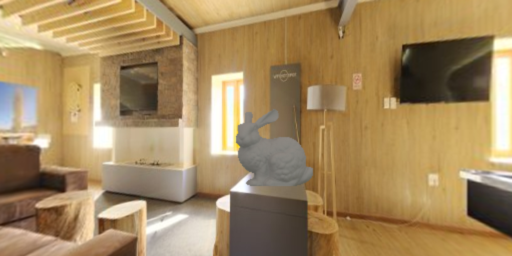}}
\hspace{0.01 cm}
\subfloat{\includegraphics[width=0.24\linewidth]{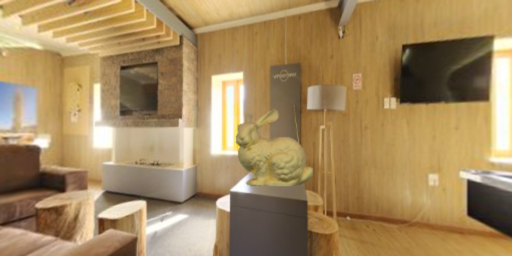}} 

\subfloat{\includegraphics[width=0.24\linewidth]{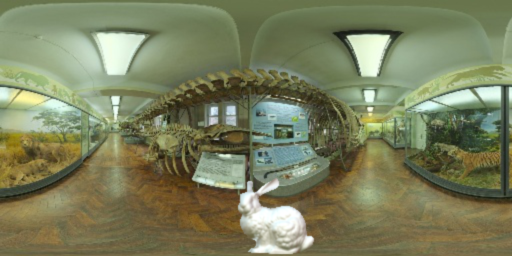}}
\hspace{0.01 cm}
\subfloat{\includegraphics[width=0.24\linewidth]{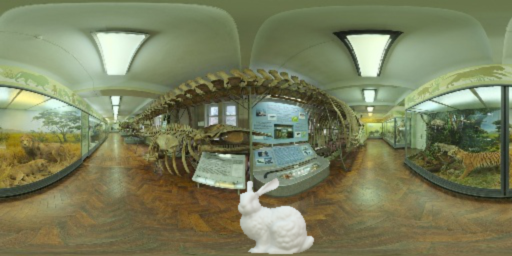}} 
\hspace{0.01 cm}
\subfloat{\includegraphics[width=0.24\linewidth]{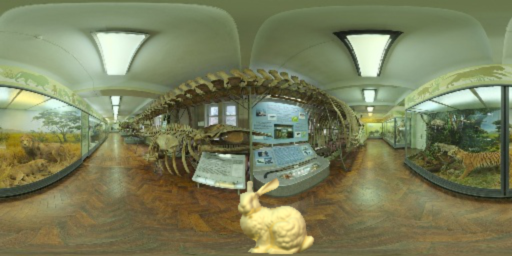}}

\subfloat{\includegraphics[width=0.24\linewidth]{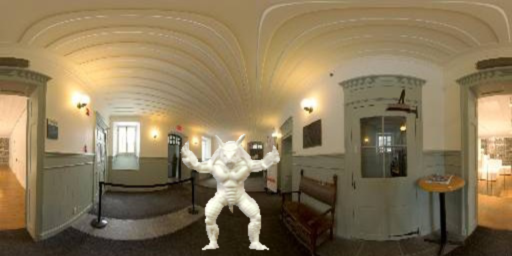}} 
\hspace{0.01 cm}
\subfloat{\includegraphics[width=0.24\linewidth]{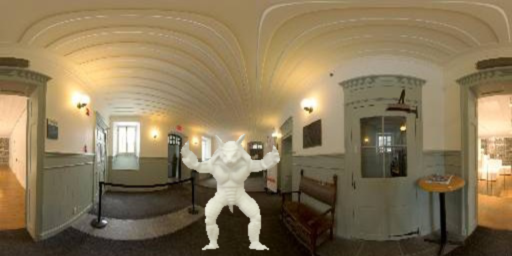}} 
\hspace{0.01 cm}
\subfloat{\includegraphics[width=0.24\linewidth]{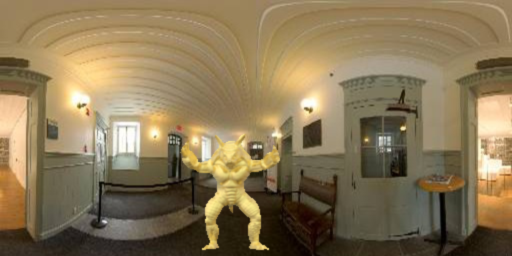}} 

\subfloat{\includegraphics[width=0.24\linewidth]{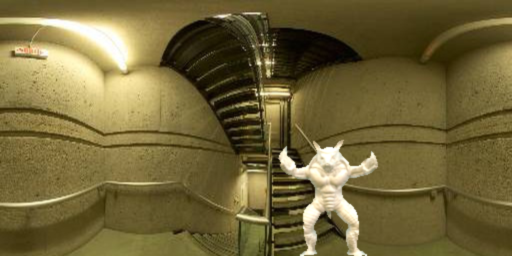}} 
\hspace{0.01 cm}
\subfloat{\includegraphics[width=0.24\linewidth]{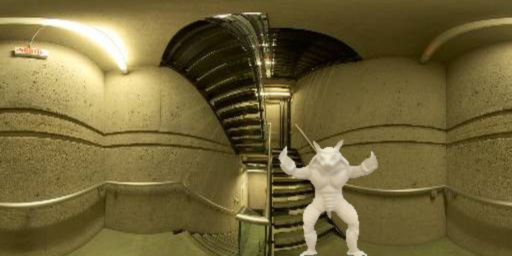}} 
\hspace{0.01 cm}
\subfloat{\includegraphics[width=0.24\linewidth]{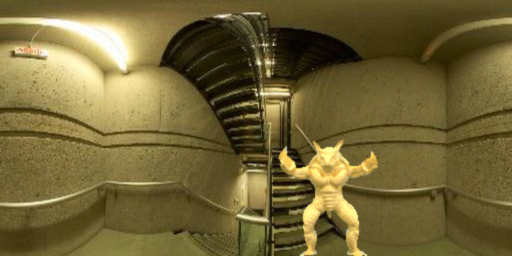}} 
    
    \caption{
    Qualitative results for virtual object rendering in real scenes with the lighting estimated by our model. 
    First seven rows are in-the-wild samples from \href{https://hdrihaven.com/}{HDRI Haven} while last two rows are from the Laval test set.
    First six examples show perspective renders (viewport denoted within the panorama), while the last
    three rows show renders within the panoramas themselves.
    Three materials are used from left to right: a conductor (reflecting mirror), rough plastic (interior and exterior index of refraction of 1.9 and 1, respectively) and another conductor (gold).
    }
    \label{fig:qualitative}
\end{figure*}

%% file: Discussion.tex
Lighting estimation is a very challenging and ill-posed problem.
Instead of relying on direct supervision, recent data-driven works opt for photometric supervision by adding spherical objects during data collection \cite{legendre2019deeplight}, relying on synthetic data \cite{garon2019fast} or on lower quality warped data \cite{song2017semantic}.
The most widely used HDR lighting dataset \cite{gardner2017learning} used in some works \cite{gardner2017learning, gardner2019deep} only contains limited samples.
Nevertheless, we showed that it is possible to learn lighting estimation in a supervised way with uncoupled data and thus, exploit the availability of high quality HDR lighting datasets in combination with larger and more diverse datasets that also contain surface information.
Further, we showed that photometric supervision through image-based relighting is sufficient to drive a lighting estimator model.
More importantly, we demonstrated that imposing the spectral distribution prior on the regressed SH coefficients greatly boosts model performance.

Nonetheless, our work only regresses up to the third order SH coefficients offering low frequency lighting environment map estimates.
While a low frequency estimation behaves good metrically, it does not capture details, which is an important goal towards realism and is left as future work.
Another issue is that without any geometrical knowledge it is not possible to embed visibility information which also detracts realistic relighting.
Most other works also predict location-conditioned outputs while we offer global estimates, offering another direction for future work.
Finally, our work is one of the first works to regress lighting from spherical panoramas, an emerging media type.

%% file: egpaper_final.bbl
\begin{thebibliography}{10}\itemsep=-1pt

\bibitem{barron2014shape}
Jonathan~T Barron and Jitendra Malik.
\newblock Shape, illumination, and reflectance from shading.
\newblock {\em IEEE transactions on pattern analysis and machine intelligence},
  37(8):1670--1687, 2014.

\bibitem{basri2003lambertian}
Ronen Basri and David~W Jacobs.
\newblock Lambertian reflectance and linear subspaces.
\newblock {\em IEEE Transactions on Pattern Analysis \& Machine Intelligence},
  (2):218--233, 2003.

\bibitem{calian2013shading}
Dan~A Calian, Kenny Mitchell, Derek Nowrouzezahrai, and Jan Kautz.
\newblock The shading probe: Fast appearance acquisition for mobile ar.
\newblock In {\em SIGGRAPH Asia 2013 Technical Briefs}, page~20. ACM, 2013.

\bibitem{chang2018matterport3d}
Angel Chang, Angela Dai, Thomas~Allen Funkhouser, Maciej Halber, Matthias
  Niebner, Manolis Savva, Shuran Song, Andy Zeng, and Yinda Zhang.
\newblock Matterport3d: Learning from rgb-d data in indoor environments.
\newblock In {\em 7th IEEE International Conference on 3D Vision, 3DV 2017},
  pages 667--676. Institute of Electrical and Electronics Engineers Inc., 2018.

\bibitem{cheng2018learning}
Dachuan Cheng, Jian Shi, Yanyun Chen, Xiaoming Deng, and Xiaopeng Zhang.
\newblock Learning scene illumination by pairwise photos from rear and front
  mobile cameras.
\newblock In {\em Computer Graphics Forum}, volume~37, pages 213--221. Wiley
  Online Library, 2018.

\bibitem{Debevec:1998:RSO:280814.280864}
Paul Debevec.
\newblock Rendering synthetic objects into real scenes: Bridging traditional
  and image-based graphics with global illumination and high dynamic range
  photography.
\newblock In {\em Proceedings of the 25th Annual Conference on Computer
  Graphics and Interactive Techniques}, SIGGRAPH '98, pages 189--198, New York,
  NY, USA, 1998. ACM.

\bibitem{debevec2004estimating}
Paul Debevec, Chris Tchou, Andrew Gardner, Tim Hawkins, Charis Poullis, Jessi
  Stumpfel, Andrew Jones, Nathaniel Yun, Per Einarsson, Therese Lundgren,
  et~al.
\newblock Estimating surface reflectance properties of a complex scene under
  captured natural illumination.
\newblock {\em Conditionally Accepted to ACM Transactions on Graphics}, 19,
  2004.

\bibitem{eigen2014depth}
David Eigen, Christian Puhrsch, and Rob Fergus.
\newblock Depth map prediction from a single image using a multi-scale deep
  network.
\newblock In {\em Advances in neural information processing systems}, pages
  2366--2374, 2014.

\bibitem{hdrcnnEilertsen2017HDRIR}
Gabriel Eilertsen, Joel Kronander, Gyorgy Denes, Rafał~K. Mantiuk, and Jonas
  Unger.
\newblock Hdr image reconstruction from a single exposure using deep cnns.
\newblock {\em ArXiv}, abs/1710.07480, 2017.

\bibitem{gardner2019deep}
Marc-Andr{\'e} Gardner, Yannick Hold-Geoffroy, Kalyan Sunkavalli, Christian
  Gagn{\'e}, and Jean-Fran{\c{c}}ois Lalonde.
\newblock Deep parametric indoor lighting estimation.
\newblock In {\em Proceedings of the IEEE International Conference on Computer
  Vision}, pages 7175--7183, 2019.

\bibitem{gardner2017learning}
Marc-Andr{\'e} Gardner, Kalyan Sunkavalli, Ersin Yumer, Xiaohui Shen, Emiliano
  Gambaretto, Christian Gagn{\'e}, and Jean-Fran{\c{c}}ois Lalonde.
\newblock Learning to predict indoor illumination from a single image.
\newblock {\em ACM Transactions on Graphics (TOG)}, 36(6):176, 2017.

\bibitem{garon2019fast}
Mathieu Garon, Kalyan Sunkavalli, Sunil Hadap, Nathan Carr, and
  Jean-Fran{\c{c}}ois Lalonde.
\newblock Fast spatially-varying indoor lighting estimation.
\newblock In {\em Proceedings of the IEEE Conference on Computer Vision and
  Pattern Recognition}, pages 6908--6917, 2019.

\bibitem{godard2017unsupervised}
Cl{\'e}ment Godard, Oisin Mac~Aodha, and Gabriel~J Brostow.
\newblock Unsupervised monocular depth estimation with left-right consistency.
\newblock In {\em Proceedings of the IEEE Conference on Computer Vision and
  Pattern Recognition}, pages 270--279, 2017.

\bibitem{green2003spherical}
Robin Green.
\newblock Spherical harmonic lighting: The gritty details.
\newblock In {\em Archives of the Game Developers Conference}, volume~56,
  page~4, 2003.

\bibitem{hold2017deep}
Yannick Hold-Geoffroy, Kalyan Sunkavalli, Sunil Hadap, Emiliano Gambaretto, and
  Jean-Fran{\c{c}}ois Lalonde.
\newblock Deep outdoor illumination estimation.
\newblock In {\em Proceedings of the IEEE Conference on Computer Vision and
  Pattern Recognition}, pages 7312--7321, 2017.

\bibitem{jakob2010mitsuba}
Wenzel Jakob.
\newblock Mitsuba renderer, 2010.

\bibitem{karakottas2019360}
Antonios Karakottas, Nikolaos Zioulis, Stamatis Samaras, Dimitrios Ataloglou,
  Vasileios Gkitsas, Dimitrios Zarpalas, and Petros Daras.
\newblock 360° surface regression with a hyper-sphere loss.
\newblock In {\em 2019 International Conference on 3D Vision (3DV)}, pages
  258--268. IEEE, 2019.

\bibitem{kingma2014adam}
Diederik~P Kingma and Jimmy Ba.
\newblock Adam: A method for stochastic optimization.
\newblock {\em arXiv preprint arXiv:1412.6980}, 2014.

\bibitem{lalonde2009estimating}
Jean-Fran{\c{c}}ois Lalonde, Alexei~A Efros, and Srinivasa~G Narasimhan.
\newblock Estimating natural illumination from a single outdoor image.
\newblock In {\em 2009 IEEE 12th International Conference on Computer Vision},
  pages 183--190. IEEE, 2009.

\bibitem{legendre2019deeplight}
Chloe LeGendre, Wan-Chun Ma, Graham Fyffe, John Flynn, Laurent Charbonnel, Jay
  Busch, and Paul Debevec.
\newblock Deeplight: Learning illumination for unconstrained mobile mixed
  reality.
\newblock In {\em Proceedings of the IEEE Conference on Computer Vision and
  Pattern Recognition}, pages 5918--5928, 2019.

\bibitem{levoy2005stanford}
Marc Levoy, J Gerth, B Curless, and K Pull.
\newblock The stanford 3d scanning repository.
\newblock {\em URL http://www-graphics. stanford. edu/data/3dscanrep}, 5, 2005.

\bibitem{li2018learning}
Zhengqin Li, Zexiang Xu, Ravi Ramamoorthi, Kalyan Sunkavalli, and Manmohan
  Chandraker.
\newblock Learning to reconstruct shape and spatially-varying reflectance from
  a single image.
\newblock In {\em SIGGRAPH Asia 2018 Technical Papers}, page 269. ACM, 2018.

\bibitem{lombardi2015reflectance}
Stephen Lombardi and Ko Nishino.
\newblock Reflectance and illumination recovery in the wild.
\newblock {\em IEEE transactions on pattern analysis and machine intelligence},
  38(1):129--141, 2015.

\bibitem{lopez2013multiple}
Jorge Lopez-Moreno, Elena Garces, Sunil Hadap, Erik Reinhard, and Diego
  Gutierrez.
\newblock Multiple light source estimation in a single image.
\newblock In {\em Computer Graphics Forum}, volume~32, pages 170--182. Wiley
  Online Library, 2013.

\bibitem{marschner1997inverse}
Stephen~R Marschner and Donald~P Greenberg.
\newblock Inverse lighting for photography.
\newblock In {\em Color and Imaging Conference}, volume 1997, pages 262--265.
  Society for Imaging Science and Technology, 1997.

\bibitem{meka2019deep}
Abhimitra Meka, Christian Haene, Rohit Pandey, Michael Zollh{\"o}fer, Sean
  Fanello, Graham Fyffe, Adarsh Kowdle, Xueming Yu, Jay Busch, Jason
  Dourgarian, et~al.
\newblock Deep reflectance fields: high-quality facial reflectance field
  inference from color gradient illumination.
\newblock {\em ACM Transactions on Graphics (TOG)}, 38(4):77, 2019.

\bibitem{NIPS2019_9015}
Adam Paszke, Sam Gross, Francisco Massa, Adam Lerer, James Bradbury, Gregory
  Chanan, Trevor Killeen, Zeming Lin, Natalia Gimelshein, Luca Antiga, Alban
  Desmaison, Andreas Kopf, Edward Yang, Zachary DeVito, Martin Raison, Alykhan
  Tejani, Sasank Chilamkurthy, Benoit Steiner, Lu Fang, Junjie Bai, and Soumith
  Chintala.
\newblock Pytorch: An imperative style, high-performance deep learning library.
\newblock In H. Wallach, H. Larochelle, A. Beygelzimer, F. d\textquotesingle
  Alch\'{e}-Buc, E. Fox, and R. Garnett, editors, {\em Advances in Neural
  Information Processing Systems 32}, pages 8024--8035. Curran Associates,
  Inc., 2019.

\bibitem{ramamoorthi2001efficient}
Ravi Ramamoorthi and Pat Hanrahan.
\newblock An efficient representation for irradiance environment maps.
\newblock In {\em Proceedings of the 28th annual conference on Computer
  graphics and interactive techniques}, pages 497--500. ACM, 2001.

\bibitem{sengupta2018sfsnet}
Soumyadip Sengupta, Angjoo Kanazawa, Carlos~D Castillo, and David~W Jacobs.
\newblock Sfsnet: Learning shape, reflectance and illuminance of facesin the
  wild'.
\newblock In {\em Proceedings of the IEEE Conference on Computer Vision and
  Pattern Recognition}, pages 6296--6305, 2018.

\bibitem{sloan2008stupid}
Peter-Pike Sloan.
\newblock Stupid spherical harmonics (sh) tricks.
\newblock In {\em Game developers conference}, volume~9, page~42. Citeseer,
  2008.

\bibitem{sloan2017deringing}
Peter-Pike Sloan.
\newblock Deringing spherical harmonics.
\newblock In {\em SIGGRAPH Asia 2017 Technical Briefs}, page~11. ACM, 2017.

\bibitem{sloan2002precomputed}
Peter-Pike Sloan, Jan Kautz, and John Snyder.
\newblock Precomputed radiance transfer for real-time rendering in dynamic,
  low-frequency lighting environments.
\newblock In {\em ACM Transactions on Graphics (TOG)}, volume~21, pages
  527--536. ACM, 2002.

\bibitem{song2019neural}
Shuran Song and Thomas Funkhouser.
\newblock Neural illumination: Lighting prediction for indoor environments.
\newblock In {\em Proceedings of the IEEE Conference on Computer Vision and
  Pattern Recognition}, pages 6918--6926, 2019.

\bibitem{song2017semantic}
Shuran Song, Fisher Yu, Andy Zeng, Angel~X Chang, Manolis Savva, and Thomas
  Funkhouser.
\newblock Semantic scene completion from a single depth image.
\newblock In {\em Proceedings of the IEEE Conference on Computer Vision and
  Pattern Recognition}, pages 1746--1754, 2017.

\bibitem{sun2019single}
Tiancheng Sun, Jonathan~T Barron, Yun-Ta Tsai, Zexiang Xu, Xueming Yu, Graham
  Fyffe, Christoph Rhemann, Jay Busch, Paul Debevec, and Ravi Ramamoorthi.
\newblock Single image portrait relighting.
\newblock {\em ACM Transactions on Graphics (TOG)}, 38(4):79, 2019.

\bibitem{weber2018learning}
Henrique Weber, Donald Pr{\'e}vost, and Jean-Fran{\c{c}}ois Lalonde.
\newblock Learning to estimate indoor lighting from 3d objects.
\newblock In {\em 2018 International Conference on 3D Vision (3DV)}, pages
  199--207. IEEE, 2018.

\bibitem{zhou2017unsupervised}
Tinghui Zhou, Matthew Brown, Noah Snavely, and David~G Lowe.
\newblock Unsupervised learning of depth and ego-motion from video.
\newblock In {\em Proceedings of the IEEE Conference on Computer Vision and
  Pattern Recognition}, pages 1851--1858, 2017.

\bibitem{zioulis2019spherical}
Nikolaos Zioulis, Antonis Karakottas, Dimitrios Zarpalas, Federico Alvarez, and
  Petros Daras.
\newblock Spherical view synthesis for self-supervised 360° depth estimation.
\newblock In {\em 2019 International Conference on 3D Vision (3DV)}, pages
  690--699. IEEE, 2019.

\end{thebibliography}
